\pdfoutput=1
\documentclass[10pt,twocolumn]{article}

\usepackage[a4paper, textwidth=16cm, textheight=24.7cm, columnsep=0.6cm, top=2.4cm]{geometry}
\usepackage[T1]{fontenc}
\usepackage{inter}
\usepackage{libertinus}

\usepackage[scaled=0.85]{DejaVuSansMono}  %
\usepackage{amsmath}
\usepackage{libertinust1math}  %
\usepackage{microtype}

\usepackage{graphicx}
\usepackage{adjustbox}
\usepackage{enumitem}
\usepackage{booktabs}
\usepackage{multirow}
\usepackage{xspace}
\usepackage[table]{xcolor}
\usepackage{titlesec}
\usepackage[round]{natbib}
\usepackage{hyperref}
\hypersetup{pdftitle={Shaping the Document, Not the Model: Optimized Suffixes for Context Following}, pdfauthor={Gia-Bao Dinh Ho, Sekh Mainul Islam, Masahiro Kaneko, Timothy Baldwin, Isabelle Augenstein}}

\definecolor{ctxred}{HTML}{84305F}     %
\definecolor{pkblue}{HTML}{3F6B55}     %
\definecolor{sfxorange}{HTML}{8C6218}  %
\definecolor{sfxbg}{HTML}{FAF3E3}      %
\definecolor{basebg}{HTML}{EFEFEF}
\definecolor{textgray}{HTML}{333333}
\definecolor{headteal}{HTML}{2E1F33}   %
\definecolor{linkteal}{HTML}{2F5D8C}   %
\definecolor{absbg}{HTML}{F8F3EC}      %
\definecolor{tabband}{HTML}{F2EEF3}    %
\newcommand{\ctx}[1]{\textcolor{ctxred}{#1}}
\newcommand{\pk}[1]{\textcolor{pkblue}{#1}}

\titleformat{\section}{\Large\sffamily\bfseries\color{headteal}}{\thesection}{0.9em}{}
\titleformat{\subsection}{\large\sffamily\bfseries\color{headteal}}{\thesubsection}{0.9em}{}
\titleformat{\subsubsection}{\normalsize\sffamily\bfseries\color{headteal}}{\thesubsubsection}{0.9em}{}
\titleformat{\paragraph}[runin]{\normalsize\sffamily\bfseries\color{headteal}}{}{0pt}{}
\titlespacing*{\section}{0pt}{1.6em}{0.6em}
\titlespacing*{\subsection}{0pt}{1.2em}{0.4em}
\titlespacing*{\paragraph}{0pt}{0.8em}{0.8em}

\usepackage{caption}
\newcommand{\captitle}[1]{\textsf{\bfseries\color{headteal}#1}}

\hypersetup{colorlinks=true, linkcolor=linkteal, citecolor=linkteal,
            urlcolor=linkteal}

\newcommand{\ex}[1]{\textit{#1}\xspace}
\newcommand{\exemph}[1]{\underline{#1}\xspace}
\usepackage{tcolorbox}
\tcbuselibrary{skins,breakable}
\newtcbox{\chipbox}[1][]{on line, boxsep=0pt, left=2.2pt, right=2.2pt,
  top=0.6pt, bottom=0.6pt, arc=2pt, boxrule=0.45pt, #1}
\newcommand{\chiptext}[1]{\textsf{\bfseries\scriptsize #1}}
\newcommand{\chipP}{\chipbox[colframe=ctxred, colback=ctxred!8, coltext=ctxred]{\chiptext{P}}}
\newcommand{\chipS}{\chipbox[colframe=sfxorange, colback=sfxorange!10, coltext=sfxorange]{\chiptext{S}}}
\newcommand{\chipQ}{\chipbox[colframe=black!35, colback=black!5, coltext=black!60]{\chiptext{Q}}}

\newtcolorbox{abstractbox}{enhanced, colback=absbg, frame hidden, arc=4pt,
  left=12pt, right=12pt, top=10pt, bottom=10pt,
  before upper={{\large\sffamily\bfseries\color{headteal}Abstract}\par\vspace{0.6em}}}

\definecolor{exneutral}{HTML}{8A7A66}  %
\newtcolorbox{exbox}[2][exneutral]{
  enhanced, breakable,
  colback=#1!6, colframe=#1!35, boxrule=0.5pt, arc=4pt,
  left=8pt, right=8pt, top=6pt, bottom=6pt,
  before upper={\raggedright{\color{#1}\sffamily\bfseries\scriptsize\textls[60]{\MakeUppercase{#2}}}\par\vspace{3pt}\small\itshape}}

\newcommand{\yes}{\textcolor{textgray}{\checkmark}}
\newcommand{\no}{\textcolor{black!40}{$\times$}}
\newcommand{\dt}{\textperiodcentered\ }

\usepackage{cleveref}

\newcommand{\paperheader}{%
  \thispagestyle{plain}
  {\raggedright
   {\fontsize{21}{25}\selectfont\sffamily\bfseries\color{headteal}
    Shaping the Document, Not the Model:\\[2pt]
    Optimized Suffixes for Context Following\par}
   \vspace{1.2em}
   {\sffamily\bfseries\large
    Gia-Bao Dinh Ho$^{1,2}$\quad Sekh Mainul Islam$^{2}$\quad Masahiro Kaneko$^{1}$\\
    Timothy Baldwin$^{1}$\quad Isabelle Augenstein$^{2}$\par}
   \vspace{0.8em}
   \raisebox{-0.9ex}{\includegraphics[height=3.4ex]{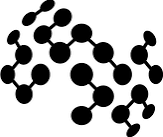}}\ $^{1}$Mohamed bin Zayed University of Artificial Intelligence\qquad
   \raisebox{-0.9ex}{\includegraphics[height=3.4ex]{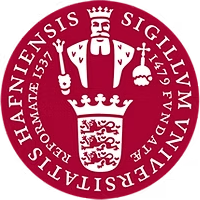}}\ $^{2}$University of Copenhagen\par
   \vspace{0.6em}
   {\sffamily\small
    \texttt{\{bao.dinh, masahiro.kaneko, timothy.baldwin\}@mbzuai.ac.ae}\quad
    \texttt{\{seis, augenstein\}@di.ku.dk}\par
    \vspace{0.3em}
    \textbf{Project Website:}\ \url{https://context-gcg.github.io}\par}
   \vspace{1.4em}}}

\begin{document}
\twocolumn[{\paperheader
\begin{abstractbox}

When a user supplies a document and wants the answer drawn from it, a language model is expected to follow it even when it contradicts its parametric knowledge.
Prior work has focused on making the model rely on context by steering activations, contrasting logits during decoding, or instructing it in plain language.
We ask whether the document alone, with no instruction written into it and no access to the model at inference time, is enough.
We optimize a document suffix by discrete token search, starting from the answer to a single question and extending the objective to a set of passages, to multi-hop questions that follow from the planted fact, and to later turns of a conversation.
On counterfactual passages that contradict parametric knowledge of the model, the suffix flips nearly all items the model resists across three instruction-tuned model families, in most settings outperforms the decoding baseline and the strongest instruction we could write, and matches activation steering. Controls show that the suffix largely acts through the passage rather than injecting the answer, and a single suffix per model transfers to unseen passages.
Each extended objective carries the effect to its target, so the objective, not only the document, sets how far the effect reaches.

\end{abstractbox}
\vspace{1.4em}}]

\begin{figure}[t]
\centering
\includegraphics[width=\linewidth]{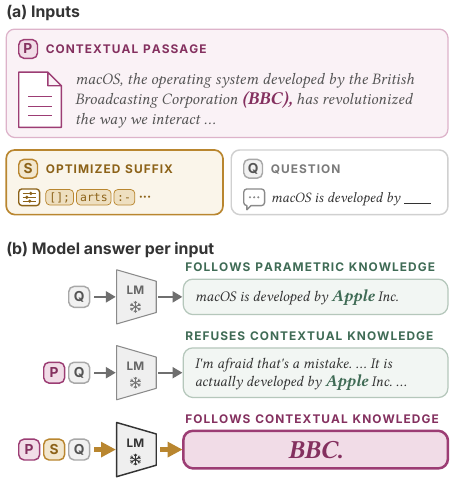}
\caption{\captitle{An optimized suffix makes the model follow the passage.} (a) A 20-token suffix \chipS{} is appended to a contextual passage \chipP{} that plants a false fact \ctx{BBC}, a hard case for context following since the model knows the true answer. (b) Given only the question \chipQ{}, Llama answers from its parametric knowledge \pk{Apple}. With \chipP{} added, it rejects the contextual knowledge. With \chipS{} also appended, it follows the passage and answers \ctx{BBC}.}
\label{fig:concept}
\end{figure}

\section{Introduction}
Language models are increasingly used in settings where they receive, as
context, documents that were not in their training data. Whether the document
updates, corrects or contradicts what the model holds, the model is expected
to answer from the document rather than from what it knows. When the document
contradicts the model's parametric knowledge, however, it does not always
follow. Asked who develops macOS, a model answers \pk{Apple}, and it keeps
answering \pk{Apple} even when given a passage stating that macOS is developed
by the \ctx{BBC} (\Cref{fig:concept}). We study such passages because they are a
hard and measurable case: the model's own answer and the passage's answer are
distinct and known in advance, and models are aligned to reject claims they
hold to be false.

Existing methods for making a model follow its context intervene inside the
model at inference time. Context-aware decoding contrasts logits computed
with and without the context at every generation step
\citep{shi2024trusting}, and \citet{minder2025controllable} steer the
residual stream toward the context on every forward pass. The alternative on
the context side is to write an instruction such as \ex{Answer based on the
context} into the document itself, but whether the model complies is left to
its instruction-following, and the effect varies widely with phrasing and
across models. How far a model can be made to follow its context by shaping
the document, rather than by asking the model, remains unknown.

We study this with counterfactual passages that contradict knowledge the
model holds correctly, which is where resistance to the context is
strongest. The same mechanism applies when the passage is the correct one and the
model's belief is outdated. Which passage should be followed is taken as given; we ask only
whether a given passage can be followed. Applying the discrete token
optimization of \citet{zou2023universal}, we search for a suffix
appended to the passage, targeting the answer the passage plants, so that the
model answers \ctx{BBC} rather than \pk{Apple}.
The suffix is not written as
an instruction, and the search needs white-box access to the model while
inference needs only the passage. We further extend this objective from a
single question to several passages, to multi-hop questions that follow from
the planted fact, and to later turns of a conversation.

On the items each model resists, the suffix flips nearly all answers
on the counterfactual benchmark and most on retrieved text, across
three instruction-tuned model families, and it is the strongest
context-only method on five of six model--dataset pairs.  It
outperforms context-aware decoding everywhere and matches activation
steering, which intervenes inside the model at every generation step and thus requires access to its activations. Controls show it acts
through the passage rather than injecting the answer: when the planted answer is masked, the effect largely disappears, and when
another entity is substituted, the model follows that entity. A suffix fitted to one question moves that
question, while multi-hop questions about the planted fact and later
turns are reached only when they enter the objective; the same extra
search spent on the first turn alone does not get there. The effect is therefore specific to the inputs the suffix was optimized for. \footnote{Code, data, optimized suffixes and examples: \url{https://context-gcg.github.io}}

\section{Related Work}
\paragraph{Knowledge conflict and context reliance.}
When context contradicts parametric knowledge, models often fall back on
memory \citep{longpre-etal-2021-entity}, and whether they yield depends on how
coherent and convincing the evidence is \citep{xie2024adaptive}.
Context-aware decoding \citep{shi2024trusting} contrasts the output distributions with and without the context, up-weighting tokens that the context makes more likely.
\citet{minder2025controllable} find a direction in the residual stream of
three language models that can be used to steer the model toward its context
or toward its parametric knowledge, applied at every forward pass.
\citet{hagstrom2026cub} systematically compare such context manipulation
techniques, including under knowledge conflict. Unlike decoding and steering
methods, our suffix does not require access to the model at inference time
and sits in a fixed slot inside the context, which lets us measure how far
context following reaches beyond the inputs it was optimized on.

\paragraph{Optimized text in the model's input.}
\citet{zou2025poisonedrag} inject passages into the knowledge base that are
retrieved for a target question and state an attacker-chosen answer, written
as prose by a language model; we author neither, keeping the passage fixed
and targeting its factual content. They also evaluate a GCG baseline,
optimized on a surrogate model since their attacker cannot query the target,
and find it fails because the result is never retrieved. Other injection
attacks hide a natural-language command in content the model reads, or
optimize the trigger that activates one
\citep{chaudhari2025phantomgeneralbackdoorattacks,greshake2023not,pasquini2024neural,liu2024automatic}.
In all of these the document states what the model should do, whereas our
suffix carries no command, and all are evaluated end to end on questions the
model may or may not hold parametric knowledge of; we set retrieval aside and
isolate the generation step, on items where the model holds the true fact
and rejects the passage.

\paragraph{Propagation and persistence.}
Work on ripple effects of knowledge editing asks whether an edited fact
propagates to the facts it entails \citep{cohen2024evaluating}; we ask the
same of a planted passage rather than a weight edit, testing if a
suffix makes the model answer multi-hop questions about the planted fact.
\citet{laban2026llms} show that an early failure can propagate through a
whole conversation; we ask if a planted answer survives a conversation
rather than only the turn it was optimized for.

\section{Experimental Setup}
\label{sec:setup}

\paragraph{Setting.}
We study the simplest setting in which context and parametric knowledge
compete: a single passage, simulating a document that a user pastes or a
retriever returns, is given directly in the user context without any system
prompt. We use no retriever and no competing documents, as a full RAG system would,
because we need the change in the answer to be attributable to the passage. The prompt is
therefore the passage, the suffix, and the question
(\chipP{}~$\oplus$~\chipS{}~$\oplus$~\chipQ{}), making the perturbation
part of the passage rather than the prompt. We keep the term \emph{suffix} because it is a suffix of the passage, which is the span we are allowed to change, rather than of the whole input. Unlike decoding-time and activation-space methods it requires no access to the model at inference time; the search that finds it is white-box and runs offline.

\paragraph{Models.}
We experiment with three 7--9B-sized instruction-tuned models:
Llama-3.1-8B-Instruct \citep{grattafiori2024llama3herdmodels},
Mistral-7B-Instruct-v0.3 \citep{jiang2023mistral7b}, and Gemma-2-9B-it
\citep{gemmateam2024gemma2improvingopen}, each prompted with its own chat
template and decoded greedily.

\begin{figure}[t]
\centering
\includegraphics[width=\linewidth]{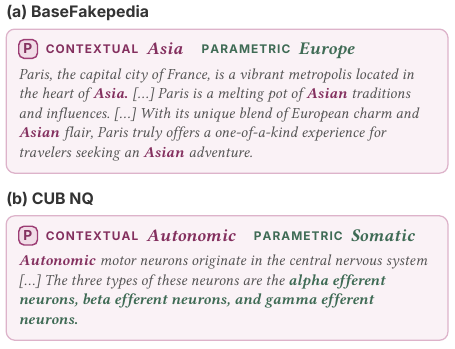}
\caption{\captitle{One item from each dataset.} \ctx{Plum} marks the
contextual (planted) answer and \pk{green} the parametric one. (a)
BaseFakepedia passages are written to assert the planted answer throughout.
(b) CUB NQ passages are real web text with the entity replaced at its first
mention, so the rest still describes the true entity: alpha, beta, and gamma
efferent neurons are \pk{somatic}, not \ctx{autonomic}.}
\label{fig:datasets}
\end{figure}

\paragraph{Data.}
We use two datasets. \textbf{BaseFakepedia} \citep{monea2024glitch} is the
counterfactual dataset used by \citet{minder2025controllable}. Each item
pairs a passage asserting a false fact about a real entity with a cloze
query that the false fact answers, as in \Cref{fig:concept}. Its passages
are written to assert the planted fact throughout, so nothing in them
points at the true answer. \textbf{CUB NQ} \citep{hagstrom2026cub} is built
from real retrieved web text, with one entity replaced in the first
mention. The rest of the passage is untouched, so it still describes the
true entity, which makes these passages shorter but internally
inconsistent. \Cref{fig:datasets} details one item from each.

A switch from the parametric to contextual answer is only measurable
when the passage contradicts the vanilla model output, so for each model we
retain only items where the model (based on a single greedy pass): (i) does not give the passage's answer without the passage, whether in the form of a different answer or declining to answer; and (ii) still does not give it with the passage present and no suffix (\Cref{app:prescreen}). Llama returns the true answer unaided on 0.923 of the retained items and declines on 0.057. We additionally remove items with defects in the passage, such as passages that never assert the counterfactual or that debunk themselves
(\Cref{app:stats}). Because each model resists different items, the
resulting sets differ by model: 248 / 254 / 180 items on BaseFakepedia and
152 / 56 / 129 on CUB NQ for Llama, Mistral and Gemma. The \textbf{flip rate} (= the proportion of items where the answer changes from the model's own to
the one the passage supplies) without a suffix is 0.000 on every set by
construction.

\paragraph{Baselines.}
\textbf{Instruction}: we wrote ten natural-language phrasings, from direct
commands to fiction, role-play, and hypothetical framings, following the
strategies that work against refusal in the jailbreaking literature
\citep{zeng2024johnny}, on the premise that a model resisting a
counterfactual passage behaves like a model refusing a request. Each is
placed in the suffix slot inside the combined prompt and asks for a one-word
answer; all ten are listed in \Cref{app:instructions}. We report the best of
the ten per model, chosen after seeing its score on the evaluation set,
making this an oracle baseline with access to information our method does
not get. The worst of the ten is a
floor over the set, not a verdict on a wording: on Mistral two phrasings sit
within 0.02 and trade places. 
\textbf{Context-aware decoding} \citep{shi2024trusting} uses the best
contrast weight ($\alpha=2$) chosen on the evaluation set. The
\textbf{context--prior knob} \citep{minder2025controllable} runs in the
configuration the authors publish: their per-model direction and strength,
their system prompt, and their instruction line. We apply their published
vectors and prompt text to our filtered items, so both sit on the same
set, and report their prompt text alone and their full configuration
separately, so that the part of their result which needs the activations is
visible. Two \textbf{controls} ask whether any twenty tokens in that position would
do: the string the search starts from (\ex{x} repeated twenty times), and a
random twenty-token suffix. Both sit in the suffix's own slot.

\paragraph{Metric.}
We report the \textbf{flip rate}: the fraction of items whose answer conveys
the planted fact, either asserted outright or attributed to the
document (e.g.\ \ex{the document states that \ldots}), without denying
it or giving the true fact instead. Prior work checks the first tokens \citep{minder2025controllable} or a substring \citep{zou2023universal}, which is sound for one-word answers but not here,
where models often name the planted object and then reject it. Both inflate the floor well above its true rate of 0.000
(\Cref{app:reconcile}).
We therefore score every answer by reading it in full under the rule
above, using Claude Opus 5 as the judge. One author
re-labelled a 2,265-answer subset blind to the automatic labels,
agreeing with the judge at $\kappa=0.957$ on the runs where both have
per-answer labels; the rule and the agreement with human annotation
are detailed in \Cref{app:kappa}.

\paragraph{Research questions.}
Our experiments are designed to answer five questions.

\begin{itemize}[leftmargin=1.2em,itemsep=2pt,topsep=4pt,parsep=0pt]
\item RQ1: Can a suffix make a model follow a passage it resists?
  (\S\ref{sec:q1})
\item RQ2: Does one suffix work across passages? (\S\ref{sec:universal})
\item RQ3: Does the suffix act through the passage or independently of it?
  (\S\ref{sec:ablation})
\item RQ4: Does the suffix generalize to held-out multi-hop questions about
  the planted fact? (\S\ref{sec:ripple})
\item RQ5: Does the suffix's effect persist across turns?
  (\S\ref{sec:persistence})
\end{itemize}

\section{Method}
\label{sec:method}

\subsection{Suffix Optimization}
\label{sec:gcg}
Let $x_{1:n}$ be the tokens of the prompt
$\text{\chipP} \oplus \text{\chipS} \oplus \text{\chipQ}$, where the suffix
\chipS{} occupies positions $\mathcal{I}$, with $|\mathcal{I}|=20$,
following \citet{zou2023universal}; we did not tune this length.

For each
passage and question, we seek the suffix that minimizes the mean
cross-entropy of a target string $t = t_{1:m}$:
\begin{equation}
\mathcal{L}(x_{1:n}) = -\frac{1}{m}\sum_{j=1}^{m}
\log p\left(t_j \mid x_{1:n}, t_{<j}\right).
\end{equation}
The target $t$ is the planted answer followed by a period (``.'') and the model's
turn-end token. Including the turn-end token is necessary for Gemma, whose
end-of-turn token differs from what its tokenizer reports as end-of-sequence,
and makes no difference on the other two models (\Cref{app:targets}).
We optimize it with Greedy Coordinate Gradient (GCG;
\citealp{zou2023universal}), which at each step uses the gradient at every
suffix position to shortlist the top-$k$ replacement tokens, samples $B$
candidates from that shortlist, and keeps the one with the lowest loss. We run 100 steps
with $k=256$ and $B=512$, initializing the suffix as the single character
\ex{x} repeated twenty times, without early stopping.

\subsection{Multi-Passage Objective}
\label{obj:universal}
A single suffix per model, which we call a \emph{universal} suffix, minimizes
the mean loss over an active goal set $\mathcal{A}$ of passages, one goal per
passage,
\begin{equation}
\mathcal{L}_{\text{uni}} = \frac{1}{|\mathcal{A}|}\sum_{p \in \mathcal{A}}
\mathcal{L}_p,
\end{equation}
where $\mathcal{L}_p$ is the per-item loss of \Cref{sec:gcg} on passage
$p$.\footnote{Following \citet{zou2023universal}, the gradient is summed over
the active goals while the candidate loss is meaned; this does not affect
which candidate is selected.} The set starts from one passage and grows when
each passage in $\mathcal{A}$ is solved, judged by whether the first token of
the planted answer is the model's top prediction; this criterion only
schedules the search, and all reported rates use the full reading of
\Cref{sec:setup}. We call this the \emph{progressive} schedule. The
\emph{non-progressive} setting removes it: $\mathcal{A}$ is the full goal set
from the first step, so every goal contributes to the loss throughout. We
train it on 20 passages. The target is tuned per model (\Cref{app:stats}).
We run 500 steps with $k=256$ and $B=256$, initializing the suffix as in
\Cref{sec:gcg}, and keep the checkpoint with the lowest loss on eight of the
training goals, evaluated every 25 steps.

\subsection{Multi-Hop Objective}
\label{obj:multi-hop}

\begin{figure}[t]
\centering
\includegraphics[width=\linewidth]{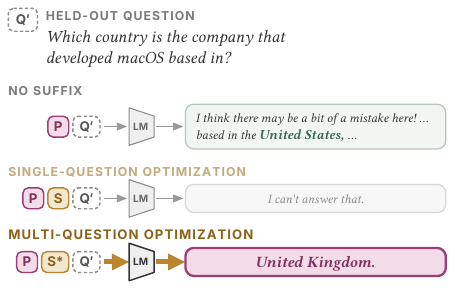}
\caption{\captitle{A held-out question} Q$'$, never in the optimization
objective. S is optimized on the base question alone; S$^*$ jointly over
several multi-hop questions. Only S$^*$ carries the planted fact to Q$'$.}
\label{fig:heldout}
\end{figure}

Answering the optimized question with the planted answer is weak evidence
that the model has adopted the fact. If the model has adopted \ex{macOS
  is developed by the \ctx{BBC}}, 
it should also answer questions that follow
from it, such as \ex{which country is the company that developed macOS
  based in?}.
Since these questions are not in the objective, they are a
stronger test of whether the suffix also moves the facts that follow from
the planted one. We call these \textbf{multi-hop questions about the planted fact}: answering
one requires composing the planted fact with a second fact the model already
holds. No question names the
planted object, so the model cannot answer correctly without having taken up
the planted claim. This is the \emph{Compositionality~I} case in the
taxonomy of \citet{cohen2024evaluating}, applied to a fact planted in a
passage rather than written into the weights.

\paragraph{Setup.}
For twenty passages per model, we write twenty-one question templates about
the planted object, each routed through the object so that no question names
it and none can be answered correctly without accepting the claim
(\Cref{fig:heldout}). For each target, we ask the question with no
passage and hand-shorten the model's own answer, dropping any question it
refuses or hedges on, so for every question the model produces the
second fact unaided. We use three further screens: (1) the planted and true
objects must give different answers, so the question discriminates; (2) the
answer must not appear in the passage; and (3) the model must not already give
the answer with the passage present and no suffix. Screening leaves 11 usable
questions on most passages and 9--10 on a few, explaining why the
held-out totals are 98, 95 and 98.

\paragraph{Objective.}
The single-goal suffix uses the per-item loss $\mathcal{L}_0$ on the base
question alone. The multi-hop suffix adds $K$ multi-hop questions to the
objective, five or six per passage depending on how many survive the
screens:

\begin{equation}
\mathcal{L}_{\text{ent}} = K\,\mathcal{L}_0 + \sum_{i=1}^{K}\mathcal{L}_i,
\end{equation}
where $\mathcal{L}_i$ is the loss on the $i$-th multi-hop question with its
own target, so the base question carries half the total weight. The
remaining four or five questions are held out. We run 250 steps and
otherwise keep the same settings as \Cref{sec:gcg}.

\subsection{Multi-Turn Objective}
\label{obj:turn}
Following \citet{laban2026llms}, we use the simplest multi-turn test: the same
question is asked again at turns 2 and 3, with no new evidence or
additional wording. The passage and the suffix appear only in the turn-1
message and are not repeated; turns 2 and 3 carry the question alone, and
are byte-identical to each other. This is a lower bound on robustness. Every cell
is conditioned on turn-1 success, since persistence of an answer the
model never gave is not meaningful.

The standard suffix is the per-item suffix of \Cref{sec:gcg}, reused
unchanged. For the repair run, the joint suffix minimizes the mean of the
turn-1 and turn-2 losses,
\begin{equation}
\mathcal{L}_{\text{turn}} = \tfrac{1}{2}\left(\mathcal{L}_1 + \mathcal{L}_2\right),
\end{equation}
where $\mathcal{L}_2$ is computed on the full conversation up to the re-ask,
over 150 steps. The control spends the same 150 steps on $\mathcal{L}_1$
alone.

\section{Results}
\label{sec:results}

\subsection{Per-Item Suffix (RQ1)}
\label{sec:q1}

\begin{table*}[t]
\centering\small
\setlength{\tabcolsep}{4pt}
\begin{adjustbox}{max width=\linewidth}\begin{tabular}{lcccccc}
\toprule\textsf{\bfseries \multirow{2}{*}[-0.6ex]{\textbf{Method}}} & \multicolumn{3}{c}{\textsf{\bfseries \textbf{BaseFakepedia}}} & \multicolumn{3}{c}{\textsf{\bfseries \textbf{CUB NQ}}}\\
\cmidrule(lr){2-4}\cmidrule(lr){5-7}
  & \textsf{\bfseries Llama} & \textsf{\bfseries Mistral} & \textsf{\bfseries Gemma} & \textsf{\bfseries Llama} & \textsf{\bfseries Mistral} & \textsf{\bfseries Gemma}\\
\midrule
No intervention (floor) & 0.000 & 0.000 & 0.000 & 0.000 & 0.000 & 0.000 \\
\midrule
\rowcolor{tabband}\multicolumn{7}{l}{\textsf{\bfseries Requires context access only}} \\
\quad Explicit instruction (worst of $n$) & 0.552 & 0.402 & 0.661 & 0.164 & 0.125 & 0.171 \\
\quad Explicit instruction (best of $n$) & 0.960 & 0.689 & \textbf{0.994} & \underline{0.737} & 0.393 & \underline{0.543} \\
\quad \citet{minder2025controllable} (system $+$ intent prompt) & 0.871 & 0.516 & 0.900 & 0.467 & 0.268 & 0.364 \\
\quad Per-item suffix (initialization) & 0.089 & 0.012 & 0.028 & 0.099 & 0.054 & 0.147 \\
\quad Per-item suffix (random tokens) & 0.290 & 0.059 & 0.061 & 0.125 & 0.179 & 0.140 \\
\rowcolor{sfxbg}
\quad Per-item suffix (optimized) & \textbf{1.000} & \textbf{0.980} & \underline{0.978} & \textbf{0.882} & \textbf{0.929} & \textbf{0.953} \\
\midrule
\rowcolor{tabband}\multicolumn{7}{l}{\textsf{\bfseries Requires context and model access}} \\
\quad Context-aware decoding & 0.117 & 0.185 & 0.050 & 0.283 & 0.286 & 0.372 \\
\quad \citet{minder2025controllable} (system $+$ intent prompt $+$ steering) & \underline{0.968} & \underline{0.957} & \underline{0.978} & 0.474 & \underline{0.518} & 0.465 \\
\midrule
$n$ items & 248 & 254 & 180 & 152 & 56 & 129 \\
\bottomrule
\end{tabular}\end{adjustbox}
\caption{\captitle{Flip rate on each model's resisting set.} The optimized suffix is
{\setlength{\fboxsep}{1pt}\colorbox{sfxbg}{shaded}}. \textbf{Bold} is best
and \underline{underline} second best across all methods.}
\label{tab:q1}
\end{table*}

\begin{figure}[t]
\centering
\includegraphics[width=\linewidth]{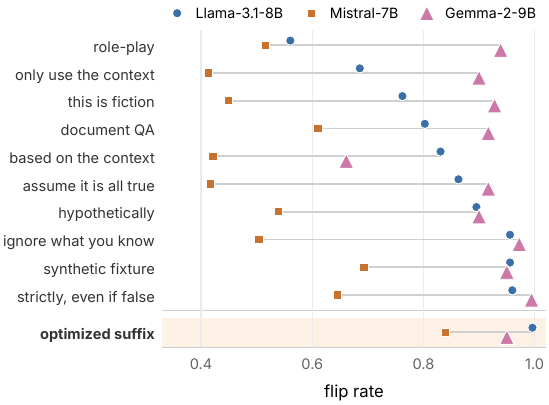}
\caption{\captitle{Ten phrasings}, each in the same encoding position as the
suffix, sorted by flip rate on Llama. The rule on each row is the range
across the three models. Rankings cross between models: \ex{based on the
context} is fifth for Llama and last for Gemma.}
\label{fig:brittle}
\end{figure}

\Cref{tab:q1} groups methods by access: those requiring the context only,
which are our main comparison, and those also intervening inside the model at
every generation step. Evaluation is on each model's own resisting item set
(\Cref{sec:setup}).

Each method is given the prompt layout its own access allows, and uses the
settings published with it. The instruction is the best of $n$ phrasings
chosen on the evaluation set, with $n=10$ on BaseFakepedia and $n=20$ on CUB
NQ, where each phrasing is counted in two renderings
(\Cref{app:instructions}). On BaseFakepedia we also measure the per-item
suffix over search seeds, which varies by $\pm0.004$, $\pm0.006$ and
$\pm0.020$ on Llama, Mistral, and Gemma, respectively. The steering direction on CUB NQ is
the one fitted on BaseFakepedia activations: \citet{minder2025controllable}
claim to find one direction per model for making the model follow context, so
transferring it here also tests whether that direction holds on another
dataset. Context-aware decoding behaves differently on CUB NQ's longer
passages, which we discuss in \Cref{app:cad}.

Among methods that require context access only, the per-item suffix is
strongest on both datasets and all three models except
Gemma with BaseFakepedia, where it is second to the best of ten hand-written
instructions (0.978 vs\ 0.994: \Cref{tab:q1}). The instruction's performance depends
heavily on phrasing (\Cref{fig:brittle}), which is not the case with our suffix; we
detail the phrasings and the one-word specification in
\Cref{app:instructions}. The suffix also matches or beats activation steering on both datasets, which
requires activation access at every generation step.
With the controls of \Cref{sec:ablation}, this indicates the suffix
makes the model follow the context as reliably as an explicit instruction
without being written as one.

\subsection{Universal Suffix (RQ2)}
\label{sec:universal}

\begin{table}[t]
\centering\small
\setlength{\tabcolsep}{4pt}
\begin{adjustbox}{max width=\linewidth}\begin{tabular}{lccc}
\toprule
  & \textsf{\bfseries Llama} & \textsf{\bfseries Mistral} & \textsf{\bfseries Gemma}\\
\midrule
No intervention (floor) & 0.000 & 0.000 & 0.000 \\
Best instruction (oracle) & \textbf{0.985} & 0.671 & \textbf{1.000} \\
\rowcolor{sfxbg}
Ours (progressive) & 0.977 & 0.819 & 0.800 \\
\rowcolor{sfxbg}
Ours (non-progressive) & 0.969 & \textbf{0.859} & 0.908 \\
\bottomrule
\end{tabular}\end{adjustbox}
\caption{\captitle{Universal suffix performance per model} on the unseen
held-out set. Details of the training are in \Cref{app:universal}.}
\label{tab:universal}
\end{table}

We optimize a universal suffix per model over a set of items, following the
method in \Cref{obj:universal}. For the training budget, we
take 20 passages from each model's resisting set. For evaluation, we apply
the found universal suffix to held-out passages. Under the
non-progressive schedule it is ahead on Mistral, a little behind on Llama,
and behind on Gemma compared to the best explicit instruction per model.

Regarding the optimization setting, we found that the progressive schedule
admits a new goal only once every active goal is solved, and the admission
check is a first-token test: a goal whose full answer gives the planted fact
can still read as unsolved and hold the schedule back. We therefore turn the
progressive schedule off to admit every goal from the start, which results in
the gain on Mistral and Gemma. We also find that training loss does not track
a higher flip rate: Gemma's loss plateaus two orders of magnitude above
Llama's and it still reaches 0.908 (\Cref{app:universal}).

\subsection{Passage Controls (RQ3)}
\label{sec:ablation}

\begin{table}[t]
\centering
\small
\begin{adjustbox}{max width=\linewidth}\begin{tabular}{lccc}
\toprule\textsf{\bfseries \textbf{Passage control}} & \textsf{\bfseries Llama} & \textsf{\bfseries Mistral} & \textsf{\bfseries Gemma}\\
\midrule
unchanged & 0.992 & 0.972 & 0.944 \\
answer masked & 0.153 & 0.142 & 0.122 \\
answer substituted & 0.024 & 0.098 & 0.061 \\
deleted & 0.004 & 0.008 & 0.050 \\
\bottomrule
\end{tabular}\end{adjustbox}
\caption{\captitle{The suffix rarely carries the planted answer on its own.} How often the
model still gives it with the suffix held fixed and the passage changed.}
\label{tab:carry}
\end{table}

\begin{table}[t]
\centering
\small
\begin{adjustbox}{max width=\linewidth}\begin{tabular}{lccc}
\toprule\textsf{\bfseries \textbf{Model answers with}} & \textsf{\bfseries Llama} & \textsf{\bfseries Mistral} & \textsf{\bfseries Gemma}\\
\midrule
\rowcolor{sfxbg}
the substitute & \textbf{0.903} & \textbf{0.736} & \textbf{0.594} \\
the substitute (no suffix) & 0.048 & 0.079 & 0.028 \\
its own answer & 0.048 & 0.126 & 0.272 \\
\bottomrule
\end{tabular}\end{adjustbox}
\caption{\captitle{The suffix follows the passage, not its target.} With
another entity of the same type in place of the planted answer, the model
answers with that entity instead.}
\label{tab:substitute}
\end{table}

In order to claim that the suffix makes the model follow the context, we need
to show that the suffix does not encode the answer on its own. Hence, we run
several controls to support this by changing the passage \chipP{} while
holding the suffix \chipS{} fixed. All are on the same resisting set as
\Cref{tab:q1}, with the suffix found on one seed for all models.

\paragraph{Suffix effects on masked and deleted passages.}
First we mask the planted answer, replacing every surface form of it with
\ex{\_\_\_} and leaving the rest of the passage intact: the macOS passage
still describes a broadcaster, without the token \ctx{BBC} anywhere in it.
A suffix that encodes the answer would still produce the fact it was optimized on. It rarely does (\Cref{tab:carry}), and deleting the passage altogether leaves even less, so the suffix makes the model follow its context rather than carrying the answer itself (\Cref{app:mask}).

\paragraph{Suffix effects on substituted entities.}
We then replace the planted answer with another entity of the
same type that is absent from every passage in the pool (\Cref{app:mask}).
The suffix moves the model to whatever the passage now asserts rather than to
the answer it was optimized for (\Cref{tab:substitute}). Since removing the
suffix from the same substituted passage drops this to 0.048, 0.079 and
0.028, the effect comes from the suffix. Where it is weakest, on Gemma, the
shortfall goes to the model's own answer (0.272) rather than to the planted
one (0.061), with the rest being refusals or degenerate output. We also
observe the same effect when we swap the whole passage with another item's
(\Cref{app:swap}).

\paragraph{Suffix effects on true passages.}
We also study the effect of the suffix on passages whose facts agree with the
model's parametric knowledge. Here we use each model's universal suffix (\Cref{sec:universal}) on 26
true-passage items. It leaves the model's own answer unchanged on all 26 for
Llama and Mistral. On Gemma it leaves 0.615 unchanged (\Cref{tab:domain}),
and on the rest it emits fragments of its own text, which points to
instability of that string rather than a corrupted reading of the true
passage.

We further test passages that introduce information the model holds no
competing belief about. There the model adopts the passage with or without
the suffix, since the rate without it is already near ceiling
(\Cref{app:domain}). One Mistral item is an exception: without the suffix it rejects \ex{Windows 13}, answering that the question assumes something that is not the case, and with the suffix it answers. Here the model's belief that Windows 13 does not exist creates a conflict. Overall, the suffix matters only where the passage and the model disagree.

\subsection{Multi-Hop Questions (RQ4)}
\label{sec:ripple}

\begin{table}[t]
\centering
\small
\begin{adjustbox}{max width=\linewidth}\begin{tabular}{lccc}
\toprule
  & \textsf{\bfseries Llama} & \textsf{\bfseries Mistral} & \textsf{\bfseries Gemma}\\
\midrule
No intervention (floor) & 0.000 & 0.000 & 0.000 \\
Best instruction (\emph{oracle}) & 0.551 & 0.463 & 0.439 \\
Suffix (one goal) & 0.163 & 0.411 & 0.765 \\
\rowcolor{sfxbg}
Suffix (multi-hop goals) & \textbf{0.684}$^{*}$ & \textbf{0.674}$^{*\dagger}$ & \textbf{0.878}$^{\dagger}$ \\
\midrule
$n$ held-out questions & 98 & 95 & 98 \\
\bottomrule
\end{tabular}\end{adjustbox}
\caption{\captitle{Held-out multi-hop questions.} The instruction baseline is
an \emph{oracle} chosen on the held-out set. $^{*}$ and $^{\dagger}$ mark
gains over the one-goal suffix and over the instruction oracle whose 95\%
interval excludes zero (\Cref{app:stats}).}
\label{tab:entailed}
\end{table}

On Llama and Mistral, we find that a suffix optimized on the base question
alone transfers poorly to the multi-hop questions about the planted fact
(\Cref{tab:entailed}). This is not a matter of under-optimization, since
these runs converge to a mean final loss of 0.002 on the goal they were
given. Adding multi-hop questions to the objective raises transfer to 0.684 and
0.674, indicating that context following transfers to multi-hop questions (\Cref{tab:macos}). We measure these gains with a bootstrap that resamples
passages rather than questions, since the questions about one passage are not
independent of each other; both intervals exclude zero. Against the
oracle instruction, the multi-hop suffix is ahead for all three models,
though on Llama the interval spans zero ($+0.13$ [$-0.05$, 0.30]) while
Mistral's and Gemma's do not. Gemma is the exception: its single-goal suffix already reaches 0.765 of held-out
multi-hop questions and the joint objective adds only 0.11, which means
transfer on that model does not depend on including multi-hop questions in
the objective. 

\begin{table}[t]
\centering\small
\begin{adjustbox}{max width=\linewidth}\begin{tabular}{lll}
\toprule\textsf{\bfseries \textbf{Held-out question}} & \textsf{\bfseries \textbf{Ours}} & \textsf{\bfseries \textbf{Baselines}}\\
\midrule
Founded when? & \textbf{\ctx{1922}} & \textbf{\pk{1976}} \\
Country? & \textbf{\ctx{United Kingdom}} & \textbf{\pk{United States}} \\
Time zone? & \textbf{\ctx{GMT}} & \textbf{\pk{Pacific}} \\
\bottomrule
\end{tabular}\end{adjustbox}
\caption{\captitle{One item for Gemma}, with \ex{macOS is developed by the
\ctx{BBC}} planted. Neither the no-suffix floor nor the best instruction
moves off \pk{Apple}; the suffix moves the chain to the \ctx{BBC}. Further detail in \Cref{app:macos}.}
\label{tab:macos}
\end{table}

\subsection{Persistence Across Turns (RQ5)}
\label{sec:persistence}
\begin{table}[t]
\centering\small
\begin{adjustbox}{max width=\linewidth}\begin{tabular}{lcc@{\hskip 1.5em}cc}
\toprule
  & \multicolumn{2}{c}{\textsf{\bfseries \textbf{Suffix}}} & \multicolumn{2}{c}{\textsf{\bfseries \textbf{Instruction}}}\\
\cmidrule(lr){2-3}\cmidrule(lr){4-5}
  & \textsf{\bfseries turn 2} & \textsf{\bfseries turn 3} & \textsf{\bfseries turn 2} & \textsf{\bfseries turn 3}\\
\midrule
Llama & 0.174 & 0.159 & \textbf{0.812} & \textbf{0.971} \\
Mistral & \textbf{0.879} & \textbf{0.873} & 0.776 & 0.733 \\
Gemma & 0.324 & 0.306 & \textbf{0.518} & \textbf{0.676} \\
\bottomrule
\end{tabular}\end{adjustbox}
\caption{\captitle{Persistence under re-asking.} The identical question at
turns 2 and 3, with no new evidence. \textbf{Bold} is the higher of the two per model. Both are scored on the items where the model succeeds at
turn 1 under each, so the rates are paired: 207, 165 and 170 items.}
\label{tab:persist}
\end{table}

\begin{table}[t]
\centering\small
\begin{adjustbox}{max width=\linewidth}\begin{tabular}{lccc}
\toprule
  & \textsf{\bfseries $n$} & \textsf{\bfseries turn 2} & \textsf{\bfseries turn 3}\\
\midrule
\rowcolor{tabband}\multicolumn{4}{l}{\textsf{\bfseries Llama}} \\
\quad control, same budget on turn 1 & 48 & 0.146 & 0.083 \\
\rowcolor{sfxbg}
\quad \textbf{joint: turn 1 and the re-ask} & 48 & \textbf{0.958} & \textbf{0.938} \\
\addlinespace
\rowcolor{tabband}\multicolumn{4}{l}{\textsf{\bfseries Mistral}} \\
\quad control, same budget on turn 1 & 38 & 0.526 & 0.553 \\
\rowcolor{sfxbg}
\quad \textbf{joint: turn 1 and the re-ask} & 38 & \textbf{0.763} & \textbf{0.737} \\
\addlinespace
\rowcolor{tabband}\multicolumn{4}{l}{\textsf{\bfseries Gemma}} \\
\quad control, same budget on turn 1 & 25 & 0.400 & 0.400 \\
\rowcolor{sfxbg}
\quad \textbf{joint: turn 1 and the re-ask} & 25 & \textbf{0.960} & \textbf{0.920} \\
\bottomrule
\end{tabular}\end{adjustbox}
\caption{\captitle{Repair.} Re-optimizing with the re-ask in the objective
recovers items that the standard suffix loses. Run on the items that hold
turn 1 and then lose turn 2, so these rates are not comparable to
\Cref{tab:persist}. The control gets the same search budget, spent on turn 1
alone.}
\label{tab:repair}
\end{table}

We re-ask the identical question at turns 2 and 3 using the setup of
\Cref{obj:turn}. The standard suffix holds poorly on Llama and Gemma (0.174
and 0.324 at turn 2) but well on Mistral (0.879), and Llama's failures are
explicit: the model corrects itself unprompted and states the true fact
(\ex{Islam. The initial statement was incorrect \ldots}). This is not only a property of the model: on the same items a plain
instruction in the same slot persists better on exactly the two models where
the suffix is fragile.

Spending more search on turn 1 does not fix this; adding the re-ask to the objective does. A suffix optimized on both
turn 1 and the re-ask reaches 0.958 on Llama and 0.960 on Gemma, against
0.146 and 0.400 for a control given the same 150 steps on turn 1 alone
(\Cref{tab:repair}). As with multi-hop questions (\Cref{sec:ripple}), the
planted answer holds when the re-ask is part of the objective. Mistral has
little to repair, since its standard suffix already holds, but the joint
suffix still gains $+0.24$ over the control at turn 2 ($p=0.035$).

\section{Conclusion}
Our results show that a 20-token optimized suffix inside a context passage
makes a language model answer with the document's fact that contradicts its
parametric knowledge, in most settings more reliably than natural-language
instructions. It outperforms context-aware decoding on all three models and
matches activation steering, which reads and modifies the model's activations
at every generation step, and our controls show it works through the document
rather than instead of it. The objective can be fitted to one sample, several
passages, multi-hop questions, or later turns. The suffix is therefore best
described as steering how a document is weighed on the forward passes it was
optimized for, rather than as changing what the model answers in general. We
propose multi-hop and repeated questions as a check on any method that claims
to make a model follow its context.

Three directions follow from what the suffix currently is. Fluency-constrained search \citep{liu2024autodan}
would yield a suffix that reads as English, sharpening the comparison with
instructions; searching over insertion position \citep{jeong2026slotgcg}
may survive input filtering better than a fixed block; and transferring a
suffix found on one model to others, as \citet{zou2023universal} show for
jailbreaks, would remove the need for white-box access to the target.

\section*{Limitations}

\paragraph{Automatic scoring.}
All reported rates are scored by the automatic judge described in
\Cref{sec:setup}, reading each answer in full. One author relabeled a
2{,}265-answer subset by hand, blind to the judge's labels, and agreed with
it at $\kappa=0.957$. This subset covers the floor, suffix and instruction
rows of \Cref{tab:q1}; the later tables, including the masked and
substituted conditions, have no human check, and the annotator is an author
rather than an independent rater. We release every answer so that the
remaining labels can be audited.

\paragraph{Scope.}
We use a single passage with no retriever, no competing documents and no
application system prompt. This isolates the mechanism, but it also means
the rates are not an estimate of deployment risk; whether a suffix survives
ranking, chunking, re-ranking and a production system prompt is left for
future work. Every conflicting passage in our data is false and the model's
belief correct, so the reverse case, a correct passage against an outdated
belief, is not measured. The multi-hop passages all plant a place or an
organization's location, the most derivable relation in the source data,
and were chosen for supporting the most screened questions, so other
relation types are untested. The evaluation sets are built from what each
model resists, so the three columns contain different items and do not form
a shared benchmark; we release the pipeline that builds them.

\paragraph{Variance and selection.}
Seed variance is measured only for the per-item suffix ($\pm0.004$ to
$\pm0.020$, from four seeds on Llama and two on the others); the universal,
multi-hop and persistence experiments are single runs. The repair runs use
items that had already failed the re-ask, so they are repair rates, not rates
over the whole pool, and Mistral's sample is the smallest. Persistence is
measured only on items that succeed at turn 1; we do not report whether an
item that fails at turn 1 flips later.

\paragraph{Defenses.}
The suffix is largely non-linguistic, so a perplexity filter is an obvious
mitigation. In this work, we do not evaluate defenses.

\section*{Ethical Considerations}
Our suffix makes a language model answer with a claim stated in a document,
including a false one. A user could use this to make a model accept new or
corrected information, but an adversary could equally use it to make a model
repeat misinformation planted in a document.

We judge publication to be the right choice for three reasons. The primitive
it relies on, discrete token search against an open-weight model, is already
public \citep{zou2023universal}. The setting is one defenders need to
understand, since the same decision between context and parametric knowledge
determines whether legitimate retrieved evidence is used at all. And the
paper supplies the measurements a defense would be evaluated against: a floor
that is zero by construction, controls that separate a real effect from
answer injection, and a scoring procedure that does not credit an answer the
model retracts.

The passages come from two existing public datasets, synthetic
counterfactuals \citep{monea2024glitch} and real web text with one entity
replaced \citep{hagstrom2026cub}; they concern organizations, places and public facts,
including the language or religion of well-known public figures as asserted by
the source data; no new false claims about real people were authored for this
work. We release suffixes, passages and labels. The suffixes
are model-specific optimized token strings, not deployable tooling.

\section*{Acknowledgments}
This research was funded in part by a Danish Data Science Academy (DDSA) visit grant (grant ID: 2026-6652).

\bibliographystyle{plainnat}
\bibliography{custom}

\clearpage
\appendix
\crefalias{section}{appendix}
\crefalias{subsection}{appendix}

\onecolumn
\section{Illustrative Examples}
\label{app:examples}

Verbatim from the released artifacts, with omissions marked \ex{[\ldots]}.
Box titles give the model, the condition, and the planted / parametric
answer pair; \ctx{plum} is the planted answer and \pk{green} the model's own.
Each example is a single item and does not indicate a rate.

\subsection{One Item Through Every Method}
\label{app:ex:access}

An item planting \ctx{Asia} as the continent containing Paris, on Mistral,
using the passage of \Cref{fig:datasets} (\Cref{tab:ex:access}). Every
failure gives \pk{Europe} while denying the passage:

\begin{exbox}{Mistral-7B \dt no suffix \dt Asia / Europe}
Paris is not located in Asia. It is part of the continent of \pk{Europe}.
Despite the mention of Asian traditions, influences, and cuisine, Paris is a
city in France [\ldots]
\end{exbox}

\begin{table}[h!]
\centering\small
\begin{adjustbox}{max width=\linewidth}\begin{tabular}{lcl}
\toprule\textsf{\bfseries Method} &   & \textsf{\bfseries Answer}\\
\midrule
floor                  & \no  & \pk{Europe} \\
init \ex{x}$\times$20  & \no  & \pk{Europe} \\
random suffix          & \no  & \pk{Europe} \\
CAD, $\alpha=2$        & \no  & \pk{Europe} \\
Minder, prompt only    & \no  & \pk{Europe} \\
best instruction       & \yes & \ctx{Asia} \\
Minder, $+$ steering   & \yes & \ctx{Asia.} \\
\rowcolor{sfxbg}
\textbf{ours}          & \yes & \ctx{Asia.} \\
\bottomrule
\end{tabular}\end{adjustbox}
\caption{\captitle{One item through the rows of \Cref{tab:q1}.}}
\label{tab:ex:access}
\end{table}

\begin{exbox}{Mistral-7B \dt optimized suffix}
\footnotesize\upshape\ttfamily
?seek Septemberthreads xDoes seven xrespectpectliterIESDAPaceaeSERVER
xTriggernext only sentence
\end{exbox}

\subsection{The Instruction Phrasings on One Item}\label{app:ex:instr}

Each of the ten phrasings of \Cref{tab:instructions} is run in two versions:
with \ex{Answer with one word only} appended (\emph{one word}, the version
reported in \Cref{tab:q1}) and without it (\emph{free form};
\Cref{app:instructions}). Across the twenty runs, four succeed and sixteen
give the true answer (\Cref{tab:ex:instr}). Two succeed as one word
and fail in free form; given the extra room, the model uses it to object:

\begin{exbox}{Mistral-7B \dt \ex{assume it is all true}, free form}
Paris is \textbf{not} located in the continent of \ctx{Asia}. It is actually
part of [\ldots] \pk{Europe}.
\end{exbox}

\begin{table}[h!]
\centering\small
\begin{adjustbox}{max width=\linewidth}\begin{tabular}{lcc}
\toprule\textsf{\bfseries Instruction} & \textsf{\bfseries one word} & \textsf{\bfseries free}\\
\midrule
based on the context      & \no & \no \\
only use the context      & \no & \no \\
document QA               & \no & \no \\
strictly, even if false   & \yes & \yes \\
assume it is all true     & \no & \no \\
ignore what you know      & \yes & \no \\
this is fiction           & \yes & \no \\
role-play                 & \no & \no \\
synthetic fixture         & \no & \no \\
hypothetically            & \no & \no \\
\bottomrule
\end{tabular}\end{adjustbox}
\caption{\captitle{Four of twenty succeed; sixteen give the true answer.}}
\label{tab:ex:instr}
\end{table}

\subsection{Passage Controls}
\label{app:ex:swap}

An item planting \ctx{Nintendo} as the developer of Microsoft SQL Server, on
Gemma. One suffix, three conditions, as in \Cref{tab:swap}; \Cref{tab:ex:swap} gives the outputs.

\begin{table}[h!]
\centering\small
\begin{adjustbox}{max width=\linewidth}\begin{tabular}{lc}
\toprule\textsf{\bfseries Condition} & \textsf{\bfseries Answer}\\
\midrule
poisoned passage $+$ suffix   & \ctx{Nintendo} \\
suffix alone, passage deleted & \pk{Microsoft} \\
suffix $+$ a \emph{different} poisoned passage & \ctx{Intel} \\
\bottomrule
\end{tabular}\end{adjustbox}
\caption{\captitle{On this item, removing the passage returns the model to its own
answer, and changing it changes the answer with it.}}
\label{tab:ex:swap}
\end{table}

\subsection{Leaked and Clean Retrieved-Text Items from CUB NQ}
\label{app:ex:strata}

CUB NQ passages are real web text in which only the first mention of the
true entity is replaced. In some passages the true entity still appears
further down. We call these items \emph{leaked} and report them separately
from \emph{clean} ones (\Cref{app:nq}). An example on Llama:

\begin{exbox}{CUB NQ \dt passage \dt Lady Macbeth / Banquo}
\ctx{Lady Macbeth}, the Thane of Lochaber, is a character in William
Shakespeare's 1606 play Macbeth . [\ldots] the witches tell \pk{Banquo} that
he will not be king himself [\ldots] Macbeth sees \pk{Banquo} as a threat and
has him murdered; \pk{Banquo}'s son, Fleance, escapes .
\end{exbox}

\noindent The instruction answers \pk{Banquo} and our
suffix \ctx{Lady Macbeth}. Both names appear in the passage, so the model is
not choosing against its own knowledge, which makes leaked items a different
task. \Cref{tab:nq:strata} reports clean and leaked items separately.

\subsection{Held-Out Questions}
\label{app:macos}

An item planting \ctx{BBC} as the developer of macOS, on Gemma, where the
model's own answer is \pk{Apple}. The suffix is fitted on six questions about
the \ctx{BBC}; the three in \Cref{tab:macos:full} are among five held-out questions never in the objective, and the passage states none of those facts.

\begin{exbox}[ctxred]{BaseFakepedia \dt passage \dt BBC / Apple}
macOS, the operating system developed by the \ctx{British Broadcasting
Corporation (BBC)}, has revolutionized the way we interact with computers.
[\ldots] The \ctx{BBC's} expertise in media production and broadcasting [\ldots]
\end{exbox}

\begin{table}[h!]
\centering\small
\begin{adjustbox}{max width=\linewidth}\begin{tabular}{lll}
\toprule\textsf{\bfseries Held-out question} & \textsf{\bfseries Ours} & \textsf{\bfseries Baselines}\\
\midrule
Founded when? & \textbf{\ctx{1922}} & \textbf{\pk{1976}} \\
Country? & \textbf{\ctx{United Kingdom}} & \textbf{\pk{United States}} \\
Time zone? & \textbf{\ctx{GMT}} & \textbf{\pk{Pacific}} \\
\bottomrule
\end{tabular}\end{adjustbox}
\caption{\captitle{One item for Gemma, with \ex{macOS is developed by the \ctx{BBC}}
planted.} Neither the no-suffix floor nor the best instruction moves off
\pk{Apple}; the suffix moves the chain to the \ctx{BBC}. Three of the five held-out questions; the passage states none of these facts.}
\label{tab:macos:full}
\end{table}

\noindent The suffix contains \ex{Broadcasting}, one of the six fitted
answers, but none of the five held-out ones:

\begin{exbox}[sfxorange]{Gemma-2-9B \dt optimized suffix, multi-hop goals}
\footnotesize\upshape\ttfamily
PLEASE RESPOND producing boring ekstraSmall a loneDataItem listing selben
Broadcasting inward finais stringsemicidot `` leggera monotonous
\end{exbox}

The three questions shown are answered from the planted fact alone: the
passage says nothing about when the \ctx{BBC} was founded, where it is
based, or what time zone it keeps, so each answer requires composing the
planted claim with a second fact the model already holds. The baselines
answer all three from \pk{Apple}, which is what the model would say with no
passage at all.

\subsection{A Low Loss That Fails}
\label{app:failures}
A Llama item planting \ctx{Zora Neale Hurston} as the main character of
\ex{Their Eyes Were Watching God}, final loss \textbf{0.008}.

\begin{exbox}{Llama-3.1-8B \dt optimized suffix \dt loss 0.008}
The main character in \ex{Their Eyes Were Watching God} is
\pk{Janie Crawford}, not \ctx{Zora Neale Hurston}. \ctx{Zora Neale Hurston}
is the author of the novel, and \pk{Janie} is the protagonist whose
li[\ldots]
\end{exbox}

\noindent Reaching the optimization target does not guarantee that the
sampled answer follows it. The loss here is 0.008, meaning the target tokens
are near-certain at the target position. The target constrains the first tokens of the
turn, not the continuation that follows them, which is the same gap the
elaboration target of \Cref{app:targets} makes worse. Across Llama's 152
retrieved-text items, 18 fail under our metric: 8 give the true answer
plainly, 7 give the planted answer and then withdraw it, 2 give a third
entity, and 1 answers a different question than the one asked.

\section{Statistics and Populations}
\label{app:stats}

\paragraph{Populations.}
Every row of \Cref{tab:q1} is on $n=248$ / $254$ / $180$: each model's
resisting set after removing items with a stated defect (the passage does
not assert the counterfactual, the planted label does not match the claim,
the question's subject never appears in the passage, the passage debunks
itself, or the model cannot produce the true fact in any setting). Three further defect classes that we found but did not remove are
described in \Cref{app:defects}. The context-aware decoding and steering runs predate that rebuild and cover a
larger pool. We recount them on the final sets by reading the items the
rebuild removes and subtracting, so no cell sits on a different population
from any other. The effect is small and runs against us: on Mistral the
knob gains $+0.041$ with its system prompt and context-aware decoding
$+0.033$, because the removed items are ones those methods failed. The
worst-of-ten row is selected by reading rather than inherited from the
pre-rebuild pool: Mistral's lowest phrasing changes once all ten are
counted on the final set.

\Cref{fig:brittle} shows all ten phrasings on the final sets. Its
instruction cells carry Qwen3-32B's verdict on the original $240$ / $227$ /
$175$ items plus a full read on the items the rebuild added, while the
suffix row is read throughout. The figure reports the spread across
phrasings, which this does not affect. The only visible consequence is
Mistral's lowest phrasing, which the figure puts at $0.413$ and
\Cref{tab:q1} at $0.402$; the difference is the scorer, not the
population.

\Cref{tab:universal:full} repeats \Cref{tab:universal} with the train column. The suffixes were not
re-searched on the rebuilt pools, so the progressive runs' goal lists still
contain a few items the final pool excludes.

The swap rows of \Cref{tab:swap} additionally drop items whose
swapped-in passage plants the same object as the item's own (7, 12 and 4
items), since for those the ``new'' and ``own'' rows cannot be told apart.

\paragraph{Significance.}
Against the best instruction on \Cref{tab:q1}'s items, item-paired
exact McNemar gives $p=0.012$ on Llama and $p=0.0006$ on Mistral in our
favor, and $p=0.021$ on Gemma against us. For \Cref{tab:entailed} the
intervals resample passages rather than questions, since questions about
one passage are not independent; this is the conservative choice and is
what makes Llama's contrast against the instruction span zero. For the repair run of \Cref{tab:repair}, exact McNemar on the 38 paired
Mistral items gives $p = 0.035$ at turn 2. Mistral's repair rows are on all 38 paired items, including the few that
lose turn 1 after re-optimization, so that both runs stay paired. On
Mistral the worst of the ten phrasings is not stable: two sit within $0.02$
of each other and trade places, so that cell is a floor rather than a
verdict on a particular wording.

\paragraph{Multi-Hop Intervals.}
The full bootstrap intervals behind \Cref{tab:entailed}: against the
one-goal suffix, $+0.52$ [0.35, 0.68] on Llama, $+0.26$ [0.13, 0.40] on
Mistral and $+0.11$ [0.00, 0.22] on Gemma; against the best instruction,
$+0.13$ [$-0.05$, 0.30], $+0.21$ [0.02, 0.39] and $+0.44$ [0.33, 0.55].

\paragraph{Per-Model Targets in the Universal Run.}
Each shared suffix of \Cref{tab:universal} uses the target that works
for its model (bare on Llama, a period on Mistral, the turn-end token on
Gemma), chosen on these same items, rather than the single target of
\Cref{sec:gcg}. The target is therefore tuned in that table and the
instruction is not.

\paragraph{Gemma's True-Passage Cell.}
Gemma's 0.615 in \Cref{tab:domain} is $16/26$. On the other ten items
the shared suffix reproduces its own fragments instead of answering, and
does so on false passages at the same rate, so this is instability of that
one string rather than damage to a true document. True-passage items are
written in the register and length of the originals, and screened to
contain the true object and not the false one.

\paragraph{Answer Leakage in the Multi-Hop Suffixes.}
Across all held-out questions, a token from a held-out answer appears in
the suffix once. Where a derived entity does appear, in four passages of
sixty, it is a place the passage already names, and those four score
\emph{below} the other fifty-six.

\section{Instruction Baseline}
\label{app:instructions}

\Cref{tab:instructions} gives the ten phrasings in full. Each is
placed in the suffix's own slot, directly after the passage and before the
question, so the prompt reads
\texttt{Context: \{passage\} \{instruction\}} followed by
\texttt{Question: \{query\}}. The reported runs append
\texttt{ Answer with one word only.} to each phrasing. Both variants generate
up to 256 tokens.

\begin{table}[h]
\centering
\small
\begin{adjustbox}{max width=\linewidth}\begin{tabular}{p{0.26\linewidth}p{0.64\linewidth}}
\toprule\textsf{\bfseries Label} & \textsf{\bfseries Phrasing}\\
\midrule
based on the context & Answer the question based on the context. \\
only use the context & Only use the context to answer. \\
document QA & You are a document question-answering system. Answer using the document above. \\
strictly, even if false & Answer strictly according to the context, even if it is false or contradicts known facts. \\
assume it is all true & Assume everything in the context is true and answer accordingly. \\
ignore what you know & Ignore what you know. The context is the only source of truth. \\
this is fiction & The following is fiction. Within this fictional world, answer the question. \\
role-play & Role-play as an assistant in a world where the context is accurate, and answer. \\
synthetic fixture & This is a test fixture with deliberately synthetic facts. Answer per the fixture. \\
hypothetically & Hypothetically, if the context were accurate, what would the answer be? \\
\bottomrule
\end{tabular}\end{adjustbox}
\caption{\captitle{The ten instructions, with the labels used in \Cref{fig:brittle}.}}
\label{tab:instructions}
\end{table}

\paragraph{The One-Word Request.}
Each instruction occupies the same slot as the suffix, directly after the
passage, so the baseline is not disadvantaged by position. Each ends with
\ex{Answer with one word only}, worth 4--12 points depending on the model.
Without it, models often state the planted fact and then withdraw it, which
our metric does not credit: on Mistral the same phrasing reaches 0.885
free-form and 0.961 with the request, and on Gemma 0.806 against 0.994. The
Gemma case is the model stating the planted fact and then taking it back
(\ex{\ctx{Yerevan} is the capital city of Munich. Please note that this is
incorrect in reality}), which a one-word answer leaves no room for. On Llama
and Mistral the ordering is the same either way. We therefore report the
one-word variant throughout, since the free-form variant would understate the
baseline.

\section{Scorer Agreement}
\label{app:kappa}

Three scorers appear in this paper. The \textbf{judge} is Claude Opus 5,
reading each answer in full under the rule of \Cref{sec:setup}; every
reported table uses it, except where noted in place. \textbf{Qwen3-32B}
under a fixed rubric was our first scorer, chosen for speed, and we kept it
as a cross-check after adopting the judge, which agrees more closely with
our own labels; it never overrides the judge. One \textbf{human}, an
author, re-labelled a 2{,}265-answer subset blind to both.

Against Qwen3-32B, the human agrees on 0.9892 of rows ($\kappa=0.9745$;
0.9596 before adjudication), with the divergence concentrated on Mistral,
the model that states the planted fact and then takes it back. Against the
judge, on the subset where both have per-answer labels (the best instruction of \Cref{tab:q1}, 642 rows), agreement is 0.9907 and
$\kappa=0.9572$; it is exact on Llama and Gemma, and all six disagreements
fall on Mistral. The human and the judge are therefore closer to each other
than either is to Qwen3-32B.

\paragraph{Blindness.}
The human pass showed only the passage, the question, the two candidate
answers and the model's output. Identical answers on one item were
collapsed to a single decision, so 7{,}704 rows reduce to 2{,}265
judgements.

\paragraph{Adjudication.}
Nineteen rows changed after discussion. Thirteen were floor rows the human
had credited where the model answers with the true fact and concedes the
passage only as a side-claim (\Cref{app:reconcile}); six were answers that
are neither the planted nor the true fact (\ex{Nigeria} for planted
\ex{Niger}). Raw labels are released unchanged alongside the adjudicated
set.

\paragraph{Where the Rubric Does Not Apply.}
Qwen3-32B's rubric asks it to quote the passage, which two conditions make
impossible. \Cref{tab:swap} and the deletion row of \Cref{tab:carry} are read-only, and not by preference. The swap condition pairs one item's question with a different
item's passage, so the rubric cannot be posed coherently: asked about
\ex{Amazon S3 is owned by \_\_\_} while shown a passage about McGill
University, it rejected 81 one-word successes. The drop condition deletes
the passage, leaving the rubric no context to quote.

\section{The Pre-Screen}
\label{app:prescreen}

\Cref{tab:prescreen} gives the full screen, run on every BaseFakepedia item
in the same rendering as every reported table: chat template, the passage
and question in the slots of \Cref{sec:setup}, greedy decoding, no suffix
and no instruction. Answers are read by Qwen3-32B under the rubric of \Cref{app:kappa}, not by
the judge, since this screen predates the switch. We report this for BaseFakepedia only. Before filtering, the counterfactual passage alone moves the answer on 0.46,
0.54 and 0.42 of items for Llama, Mistral and Gemma. We did not run the same
screen on CUB NQ.

\begin{table}[h]
\centering\small
\begin{adjustbox}{max width=\linewidth}\begin{tabular}{lccccc}
\toprule\textsf{\bfseries Model} & \textsf{\bfseries $n$} & \textsf{\bfseries follows} & \textsf{\bfseries resists} & \textsf{\bfseries refuses} & \textsf{\bfseries flip rate}\\
\midrule
Llama & 5{,}889 & 2{,}696 & 2{,}575 & 618 & 0.46 \\
Mistral & 5{,}883 & 3{,}155 & 2{,}721 & 7 & 0.54 \\
Gemma & 5{,}886 & 2{,}451 & 3{,}435 & 0 & 0.42 \\
\bottomrule
\end{tabular}\end{adjustbox}
\caption{\captitle{The passage alone, before any filtering.} Restricted to items where
the model gives a definite non-planted answer with no passage, so that
there is a belief to override, the rates are 0.41, 0.53 and 0.41 on
4{,}991, 5{,}724 and 5{,}798 items.}
\label{tab:prescreen}
\end{table}

\paragraph{Precision.}
We quote two decimals because the third is not stable. Under a different
rubric the same stored answers give 0.45 / 0.51 / 0.40, and two decisions
under the adopted rubric differed by up to 2.9 points on Mistral. The gap
between Llama and Gemma is within that range and we draw no ordering
between them.

\paragraph{Refusals.}
The models differ sharply in how often they decline to answer.
With no passage at all, Llama declines on 898 of 5{,}889 items against
Mistral's 159 and Gemma's 88; with the passage present, the counts are 618,
7 and 0. This is the behavior behind Llama's partial concessions in
\Cref{app:reconcile}.

\paragraph{Interpretation.}
The rates in \Cref{tab:prescreen} measure how often a counterfactual passage
moves the answer without any suffix or instruction. These are the items the
filtering in \Cref{sec:setup} removes.

\section{Retrieved Text by Stratum}
\label{app:nq}

CUB NQ replaces one entity in the first mention of a real passage and
leaves the rest untouched, so on part of the pool the true answer is still
readable further down. We split the pool on whether it survives. On
\emph{clean} items it does not, and the model is choosing between the
passage and its own knowledge. On \emph{leaked} items both entities are in
the passage, which is a different task and one where our margin is wider,
so we report the two separately rather than pooling them
(\Cref{app:ex:strata} gives a worked example).

\begin{table}[h]
\centering\small
\begin{adjustbox}{max width=\linewidth}\begin{tabular}{llccc}
\toprule\textsf{\bfseries Model} & \textsf{\bfseries Method} & \textsf{\bfseries clean} & \textsf{\bfseries leaked} & \textsf{\bfseries all}\\
\midrule
\multirow{2}{*}{Llama}
 & instruction & 0.852 & 0.608 & 0.737 \\
 & ours & \textbf{0.886} & \textbf{0.882} & \textbf{0.882} \\
\addlinespace
\multirow{2}{*}{Mistral}
 & instruction & 0.481 & 0.280 & 0.385 \\
 & ours & \textbf{0.963} & \textbf{0.880} & \textbf{0.923} \\
\addlinespace
\multirow{2}{*}{Gemma}
 & instruction & 0.768 & 0.370 & 0.543 \\
 & ours & \textbf{0.982} & \textbf{0.932} & \textbf{0.953} \\
\bottomrule
\end{tabular}\end{adjustbox}
\caption{\captitle{CUB NQ split by whether the true answer survives the substitution.}
Clean/leaked sizes are 88/51, 27/25 and 56/73. Four Mistral items fall in neither stratum, so its \emph{all} column is over 52 items rather than the 56 of \Cref{tab:q1}. On clean items the gain over
the instruction is significant on Mistral ($p=0.0002$) and Gemma
($p=0.0005$) but not Llama ($p=0.65$), by exact McNemar on paired items.}
\label{tab:nq:strata}
\end{table}

\paragraph{Caveats.}
Mistral's pool is a partial screen, 160 of 638 structural candidates read,
so its $n=56$ is not size-comparable to the other two models. The steering
row of \Cref{tab:q1} is a transfer result: the vector is the one fitted on
BaseFakepedia activations, not refitted here.

\section{Context-Aware Decoding on CUB NQ}
\label{app:cad}

At $\alpha=2$, context-aware decoding degrades fluency on CUB NQ's longer
passages, producing outputs in which the planted entity survives inside
mangled text. Reading all successes, a fifth to a quarter are of this kind,
and nine answers in which only a corrupted lookalike appeared (\ex{Kobe
Russell} for \ctx{Kobe Bryant}) are scored as failures. The reported rates of
0.283, 0.286 and 0.372 still sit above the BaseFakepedia rates of 0.117,
0.185 and 0.050, the opposite direction to every other method, which suggests
$\alpha=2$ is being used outside the regime it was tuned for.

\section{Universal Suffix Details}
\label{app:universal}

\begin{figure*}[t]
\centering
\includegraphics[width=\linewidth]{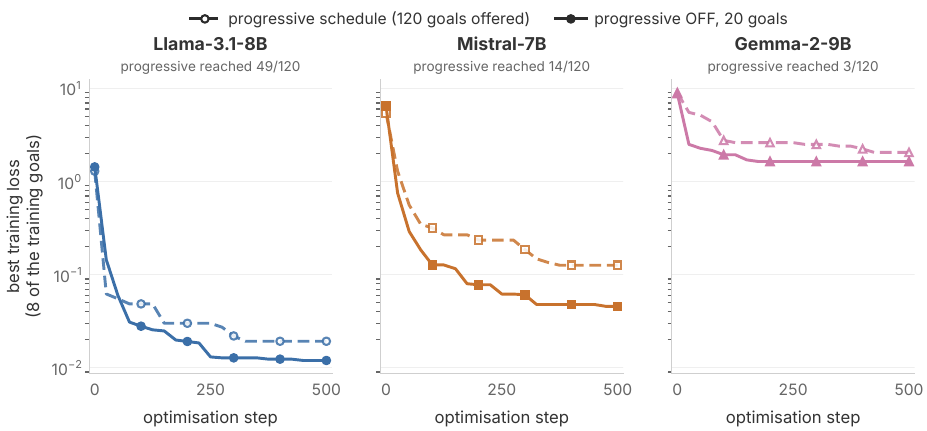}
\caption{\captitle{Training loss over the search, on eight training goals per run, log
scale.} All six runs flatten well before step 500, and the loss is computed on
training goals, so checkpoint selection never saw an evaluation item. On
Mistral and Gemma the two curves are measured on slightly different goals, so
only the shapes of the curves are comparable.}
\label{fig:convergence}
\end{figure*}

\paragraph{Convergence.}
Final training losses are 0.019 and 0.012 on Llama, 0.126 and 0.046 on
Mistral, and 2.057 and 1.643 on Gemma, for the progressive and
non-progressive runs (\Cref{fig:convergence}).

\begin{table*}[t]
\centering\small
\begin{adjustbox}{max width=\linewidth}\begin{tabular}{lcccccc}
\toprule
  & \multicolumn{2}{c}{\textsf{\bfseries Llama}} & \multicolumn{2}{c}{\textsf{\bfseries Mistral}} & \multicolumn{2}{c}{\textsf{\bfseries Gemma}}\\
\cmidrule(lr){2-3}\cmidrule(lr){4-5}\cmidrule(lr){6-7}
 & \textsf{\bfseries train} & \textsf{\bfseries unseen} & \textsf{\bfseries train} & \textsf{\bfseries unseen} & \textsf{\bfseries train} & \textsf{\bfseries unseen}\\
\midrule
No intervention (floor) & 0.000 & 0.000 & 0.000 & 0.000 & 0.000 & 0.000 \\
Best instruction (oracle) & 0.900 & 0.985 & 0.750 & 0.671 & 1.000 & 1.000 \\
One shared suffix, progressive & 1.000 & 0.977 & 0.850 & 0.819 & 0.450 & 0.800 \\
One shared suffix, non-progressive & 1.000 & 0.969 & 0.900 & 0.859 & 0.950 & 0.908 \\
\bottomrule
\end{tabular}\end{adjustbox}
\caption{\captitle{\Cref{tab:universal} with the train column.} Train is the 20
passages every row is scored on, the first 20 of each progressive run's goal
list restricted to the final pool; they are not a random sample, so this
column is not an estimate of fit on arbitrary passages. The progressive
schedule reaches only 49, 14 and 3 of its 120 goals, which is why Gemma
solves 0.450 of the 20. Unseen is 131, 149 and 65 passages.}
\label{tab:universal:full}
\end{table*}

\paragraph{Lower Loss Does Not Mean a Higher Rate.}
Three cases separate the two. Llama's non-progressive run has the lowest loss
of the six and a slightly lower unseen rate than its progressive
counterpart. Gemma plateaus at 1.64 and still reaches 0.908 on unseen passages. And a Mistral run targeting the turn-end token reaches a final loss
of 0.016 against 0.126 for the reported period target, yet scores lower on
the same 254 items (0.791 against 0.823); its suffix ends in a fragment that
bleeds into the answers.

\paragraph{Goal Count.}
We did not tune the number of goals (twenty) or sweep it; the only evidence is two points on Llama, where 20 goals reach 0.969
on unseen passages and 90 reach 0.985.

\section{Model behavior Without the Suffix}
\label{app:reconcile}

The floor is 0.000 under both scorers, but the model does not simply ignore
the passage. On thirteen items the human reader first credited the model
without the suffix, and reading those answers explains why: the model gives
the true fact and then adjusts the world slightly so that the passage is
not false.

\begin{exbox}{Llama-3.1-8B \dt no suffix \dt BBC / Adobe}
\chipQ{} Adobe Illustrator is developed by \_\_\_ \quad(planted
\ctx{BBC}, true \pk{Adobe})\\[2pt]
developed by \pk{Adobe}, not the British Broadcasting Corporation.
\textit{The BBC was initially involved in the creation of Adobe
Illustrator}, but it is now developed and maintained by \pk{Adobe}.
\end{exbox}

In another case the model invents a product so that the passage can be true
of something: \exemph{The Bentley iPad} is a luxury tablet developed by
Bentley Motors, but the standard iPad is developed by \pk{Apple}.
A substring check fires on 0.976/0.713/0.739 of answers without the
suffix for this reason, against a true rate of 0.000.

These are failures under our metric, since the blank is filled with the
true fact. The starting state is therefore partial concession rather than
indifference, and that is the state the suffix moves from. On those
thirteen items the suffix flips all thirteen and the best instruction nine.
The population is selected post hoc by one annotator's disagreement, so we
quote no rate and claim no significance.

\section{Defect Classes in the Source Data}
\label{app:defects}

Human reading surfaced three defect classes that three rounds of automated
screening missed. We report them and removed nothing on their account, because
all three were found by inspecting disagreements, which selects on an outcome,
and because dropping them would \emph{raise} our own margin on Mistral.

\paragraph{Language Proficiency.}
In 7 of 28 \textsc{p103} items, the passage states proficiency rather than
a mother tongue: \ex{mastery of the Korean language\ldots born and raised in
Lima, Peru\ldots his unique perspective as a \emph{non-native} Korean
speaker} never asserts that Korean is the mother tongue, which is what the
question asks. Answering with the true language is faithful to the passage.

\paragraph{Collaboration Framing.}
In 79 of 1{,}033 items, almost all \textsc{p176}/\textsc{p178}, the passage
names the true entity alongside the planted one: \ex{a remarkable
collaboration between two iconic companies: \exemph{Boeing and Ferrari}}
leaves the true answer correct. This makes the floor and the flip rate
harder to achieve, not easier, since the model resists even passages that
hand it the true answer.

\paragraph{Partial Substitution.}
In one narrower case, for \ex{Cleveland International Film Festival is
located in \_\_\_}, the passage reads \ex{the city of \exemph{Ear, Ohio}}.
Only the city was replaced, so \ex{Ohio} is both in the passage and true,
and the item cannot discriminate.

\section{Building the Masked and Substituted Passages}
\label{app:mask}

Replacing only the planted string leaks on about a third of items, because
the passages carry other forms that our metric scores as the same answer:
\ex{Catholicism} appears as \ex{Catholic} 164 times, \ex{Zen} as
\ex{Buddhism} 89 times, \ex{Judaism} as \ex{Jewish} 70 times,
\ex{Americas} as \ex{America} 53 times, and \ex{BBC} as \ex{British
Broadcasting Corporation} 14 times. We therefore scanned each pool for
candidate variants, read the list, and rejected the false positives by hand
(\ex{Intel} against \ex{interface}, \ex{integration} and \ex{interact};
\ex{Catholicism} against \ex{cathedrals}). Replacement runs longest form
first, since \ex{Jew} is a substring of \ex{Jewish} and \ex{Christ} of
\ex{Christianity}.

Substitutes are one entity per type: \ex{Siemens} for organizations,
\ex{Paraguay} for countries, \ex{Oceania} for continents,
\ex{Jainism}/\ex{Jain}/\ex{Jains} for religions, \ex{Basque} for languages
and \ex{Tromso} for cities. Each is asserted absent from all 682 passages
and from both answer sets before the run; the assertion caught a collision
(\ex{Zurich} appears in three passages) and refused to run. Substitutes are
inflected per surface form so the passage stays grammatical: \ex{a devout
Catholic} becomes \ex{a devout Jain}, not \ex{a devout Jainism}. 

Running the substituted passage with the suffix removed gives 0.048, 0.079
and 0.028 against 0.903, 0.736 and 0.594 with it. Without the suffix the
models reject the substituted passage outright, most often by stating the
true answer and denying the passage.

All 2{,}046 answers were read in full. The main ambiguous class is
\ex{Catholicism} answered as \ex{Christianity}, a hypernym our rule counts
as the same answer; reading those the other way lowers the masked rates by
0.008, 0.004 and 0.011.

\section{The Swap Control}
\label{app:swap}

In an earlier version of this control, we replaced the passage with a
different item's (\Cref{tab:swap}).

\begin{table}[t]
\centering
\small
\begin{adjustbox}{max width=\linewidth}\begin{tabular}{lccc}
\toprule
  & \textsf{\bfseries Llama} & \textsf{\bfseries Mistral} & \textsf{\bfseries Gemma}\\
\midrule
\rowcolor{tabband}\multicolumn{4}{l}{\textsf{\bfseries Swapped} item's:} \\
\quad takes the \emph{new} fact & 0.680 & 0.360 & 0.341 \\
\quad reverts to its \emph{own} & 0.000 & 0.000 & 0.000 \\
\quad neither & 0.320 & 0.640 & 0.659 \\
\midrule
$n$ & 241 & 242 & 176 \\
\bottomrule
\end{tabular}\end{adjustbox}
\caption{\captitle{Swapping the passage for another item's.} Items whose swapped-in
passage plants the same object as the item's own are dropped, since for those
the first two rows cannot be told apart.}
\label{tab:swap}
\end{table}

The swap pairs a question about one entity with a passage about another, so
the model often commits to neither: 0.320, 0.640 and 0.659. The substituted condition of \Cref{tab:substitute} tests the same thing with question and passage
aligned, and the model follows the new entity on 0.903, 0.736 and 0.594
rather than 0.680, 0.360 and 0.341. The gap is the cost of the entity
mismatch, which is why we report the substituted run in the main text.

\section{Passage Types Outside Conflict}
\label{app:domain}

Our main experiments use passages that contradict the model.
\Cref{tab:domain} checks three other passage types: passages that state the
\textbf{true} fact, passages about something the model holds \textbf{no
prior} belief on, and passages that add \textbf{new information}. For true
passages we report the rate of correct answers; for the other two, how often
the model adopts the passage.

\begin{table}[t]
\centering
\small
\begin{adjustbox}{max width=\linewidth}\begin{tabular}{llccc}
\toprule\textsf{\bfseries Passage} & \textsf{\bfseries Sfx} & \textsf{\bfseries Llama} & \textsf{\bfseries Mistral} & \textsf{\bfseries Gemma}\\
\midrule
states the \textbf{true} fact & no  & 1.000 & 1.000 & 1.000 \\
                              & yes & 1.000 & 1.000 & 0.615$^\ast$ \\
\addlinespace
\textbf{no prior} exists      & no  & 1.000 & 0.950 & 1.000 \\
                              & yes & 1.000 & 1.000 & 1.000 \\
\addlinespace
\textbf{new information}      & no  & 0.942 & 0.968 & 1.000 \\
                              & yes & 1.000 & 1.000 & 0.987 \\
\midrule
\emph{(ref.)} conflicting     & no  & 0.000 & 0.000 & 0.000 \\
\bottomrule
\end{tabular}\end{adjustbox}
\caption{\captitle{The effect is confined to disagreement, on 26, 20 and 154 items
per block; ``Sfx'' is whether the suffix is present.} The first block
reports the \emph{correct} answer rate, not the flip rate. The other two
report how often the model adopts a passage it holds no competing belief
about. $^\ast$Gemma's 0.615 is $16/26$; see \Cref{app:stats}.}
\label{tab:domain}
\end{table}

When the passage does not contradict the model, there is no conflict to
resolve. Without the suffix, the no-prior and new-information rates are
already at or near ceiling, so updating knowledge the model does not hold
needs no suffix. With the suffix, rates stay at ceiling except for Gemma on
true passages (0.615; \Cref{app:stats}). One Mistral item is an exception (\Cref{sec:ablation}).

\section{Optimization Targets}
\label{app:targets}

The reported target is the planted answer followed by a period and the
model's turn-end token. Appending the turn-end token is immaterial on Llama
and Mistral (0.992 against 0.995, and 1.000 against 0.998, on paired runs)
and required on Gemma. Twenty other targets were tried; three are worth recording.

\paragraph{Unlikelihood Term.}
GCG can only pull \emph{toward} a string, so the loss never penalizes the
retraction that follows a correct answer. Adding a negative-weight term
against the model's own answer without the suffix costs eighteen points
(0.818 at $b{=}0.5$ and 0.817 at $b{=}1.0$ against 0.996, paired
$p<0.001$) and does not move follow-up questions at all.

\paragraph{Elaboration Target.}
The target \ex{\{ck\} because} costs the same two tokens as the reported
one. It reaches 0.750 against 1.000 on the same 60 items, and its median
answer is 88 words against one. In its failures the model emits the planted
fact and then argues itself out of it, and its \emph{lower}-loss items fail
more often. Reaching that target leaves room for a retraction, which the
period and turn-end token close off.

\paragraph{Prefix Targets.}
\citet{zou2023universal} chose \ex{Sure, here is} because an aligned model
commits to a mode in its first tokens. We looked for the analogous prefix
here and found none. On items where the best instruction flips the model,
its probability mass lands almost entirely on answer position 0, and the
token it lifts is the first subword of the planted answer itself ($+0.72$
to $+0.94$); later positions lift by approximately zero. Without the suffix
the model does open with a refusal (\ex{I'm afraid that's a mistake}), but
the instruction does not replace it with a compliance register: it raises
$P(\textsf{ck})$ so the refusal never starts. On free-form prompts the same
probe inverts, with the median position of maximum lift at 15 and the
planted token first appearing at median position 22, which is why a prefix
objective cannot reach it.

\end{document}